\documentclass{article}
\usepackage{spconf,amsmath,graphicx}
\usepackage{amsfonts}
\usepackage[table,dvipsnames]{xcolor}
\newcolumntype{L}{@{}>{\kern\tabcolsep}l<{\kern\tabcolsep}}
\usepackage{booktabs}
\usepackage{multirow}
\usepackage{enumitem}
\usepackage{textpos}

\usepackage[pages=some,placement=top]{background}

\DeclareMathOperator*{\median}{\textit{med}}

\definecolor{pitchblack}{cmyk}{1 1 1 1}

\title{TUMORNET: Lung nodule characterization using multi-view  Convolutional Neural Network with Gaussian Process}
\name{Sarfaraz Hussein$^1$, Robert Gillies$^2$, Kunlin Cao$^3$, Qi Song$^3$, Ulas Bagci$^1$}
\address{$^1$Center for Research in Computer Vision (CRCV) at University of Central Florida, Orlando, FL. \\ 
$^2$H. Lee Moffitt Cancer Center and Research Institute, Tampa, FL.\\
$^3$CuraCloud Corporation, Seattle, WA.}

\begin{document}
%
\backgroundsetup{contents=Accepted for publication in IEEE International Symposium on Biomedical Imaging (ISBI) 2017,color=black!100,scale=1.5,position={5.5,1.25}}
\BgThispage

\maketitle

\begin{abstract}
Characterization of lung nodules as benign or malignant is one of the most important tasks in lung cancer diagnosis, staging and treatment planning. While the variation in the appearance of the nodules remains large, there is a need for a fast and robust computer aided system. In this work, we propose an end-to-end trainable multi-view deep Convolutional Neural Network (CNN) for nodule characterization. First, we use median intensity projection to obtain a 2D patch corresponding to each dimension. The three images are then concatenated to form a tensor, where the images serve as different channels of the input image. In order to increase the number of training samples, we perform data augmentation by scaling, rotating and adding noise to the input image. The trained network is used to extract features from the input image followed by a Gaussian Process (GP) regression to obtain the malignancy score. We also empirically establish the significance of different high level nodule attributes such as calcification, sphericity and others for malignancy determination. These attributes are found to be complementary to the deep multi-view CNN features and a significant improvement over other methods is obtained.
\end{abstract}
\section{Introduction}
\label{sec:intro}

Lung cancer accounts for the highest number of mortalities among all cancers in the world. Classification of lung nodules into malignant and benign is one of the most important tasks in this regard. A fast, robust and accurate system to address this challenge would not only save a lot of radiologists' time and effort, but would also enable the discovery of new discriminative imaging features. Significant successes in terms of improved survival rates for lung cancer patients have been observed due to improvements in CAD (Computer Aided Diagnosis) technologies and development of advanced treatment options. However, lung cancer still has a 5-year survival rate of 17.8\% where only 15\% of all cases are diagnosed at an early stage \cite{cancerstats}.

Conventionally, the classification of lung nodules was performed using hand-crafted imaging features such as histograms \cite{uchiyama2003quantitative}, Scale Invariant Feature Transform (SIFT) \cite{farag2011evaluation}, Local Binary Patterns (LBP) \cite{sorensen2010quantitative} and Histogram of Oriented Gradients (HOG) \cite{song2013feature}. The extracted sets of features were then classified using a variety of classifiers such as Support Vector Machines (SVM) \cite{orozco2015automated} and Random Forests (RF) \cite{ma2016automatic}. Recently with the success of deep convolutional neural network (CNN) for image classification, the detection and classification applications in medical imaging have adopted it for improved feature learning and representation. Tedious feature extraction and selection can now be circumvented using supervised high level feature learning. This has also attracted the attention of researchers working in lung nodule detection and classification with limited success since the feature learning and classification were considered as separate modules. In those frameworks a \textit{pre-trained} CNN was only used for feature extraction whereas classification was based on an off-the-shelf classifier such as RF \cite{kumar2015lung,buty2016characterization}. In sharp contrast to these methods, we perform an end-to-end training of CNN for nodule characterization while combining multi-view features to obtain improved characterization performance. \\

\noindent \textbf{Contributions:}
\begin{itemize}[leftmargin=*]
\itemsep0em 
\item We perform an end-to-end training of CNN from scratch in order to realize the full potential of the neural network i.e. to learn discriminative features. 
\item Extensive experimental evaluations are performed on a dataset comprising lung nodules from more than 1000 low dose CT scans.
\item We empirically establish the complementary significance of high level nodule attributes such as calcification, lobulation, sphericity and others along with the deep CNN features to improve the malignancy determination. 
\end{itemize}

\begin{figure*}[h]
\hspace{-0.1in}
\includegraphics[width=180mm]{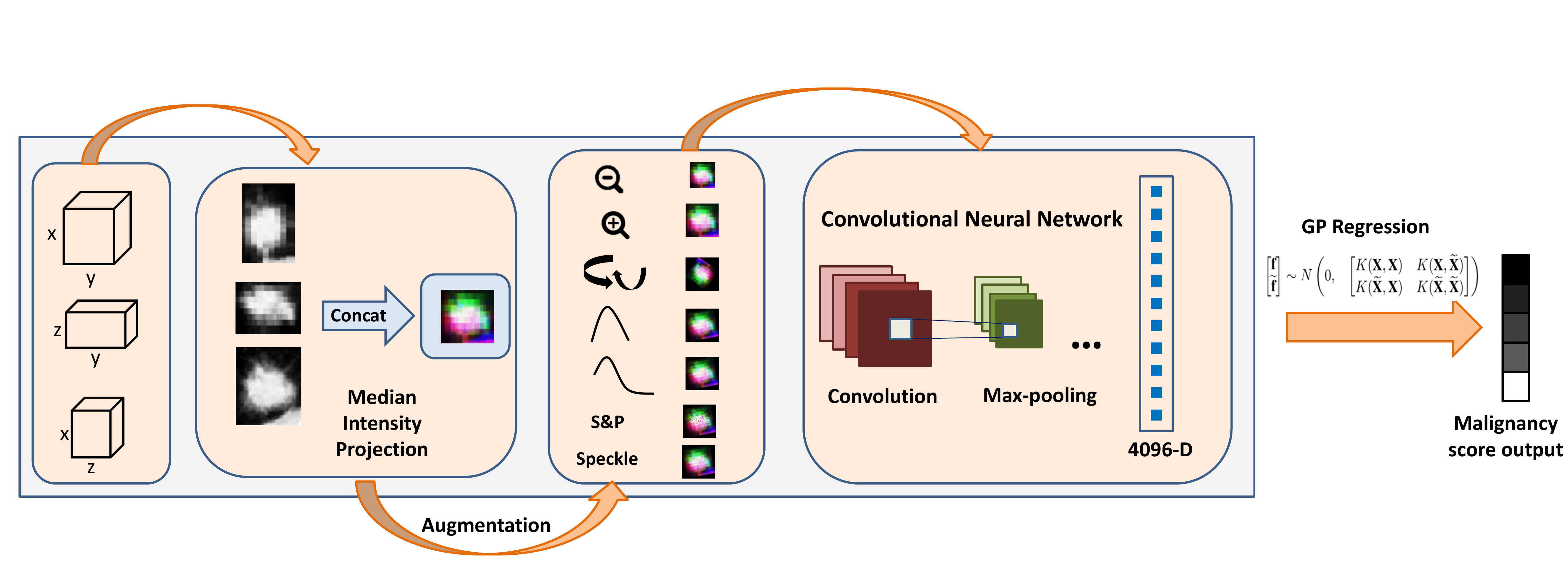}\vspace{-0.4 cm}
\caption{An overview of the proposed method. First, the median intensity projection is performed across each axis, followed by their concatenation as three channels of an image. Data augmentation is performed using scaling, rotation, adding Gaussian, Poisson, Salt and Pepper (S\&P) and Speckle Noise. A CNN with 5 convolution and 3 fully connected layers is trained from scratch. For testing, the 3 channel image is passed through the trained network to get a 4096 dimensional feature from the first fully connected layer. Finally, the malignancy score is obtained using the Gaussian Process regression.}
\label{fig:workflow}
\vspace{-0.4 cm}
\end{figure*}

\section{Materials}
\label{sec:format}

We evaluated our proposed approach on LIDC-IDRI dataset from Lung Image Database Consortium \cite{armato2011lung}, which is one of the largest publicly available lung cancer screening dataset. The dataset comprises 1018 scans with slice thickness varying from 0.45 mm to 5.0 mm. Four expert radiologists annotated lung nodules with diameters greater than or equal to 3 mm. In our training and evaluation framework, we sampled nodules which were annotated by at least three radiologists. There were 1340 nodules satisfying this criterion. The nodules have a malignancy rating from 1 to 5 where 1 represents low malignancy and 5 is for highly malignant nodules. We excluded nodules with an average score equal to 3 to account for uncertainty in the decision of the radiologists. Our final dataset consists of 635 benign and 510 malignant nodules for classification. All images were resampled to have 0.5 mm spacing in each dimension.
\vspace*{-3ex}
\section{Method}
\label{sec:pagestyle}
\subsection{Multiview Deep Convolutional Neural Network:}
\textbf{Architecture:}\\
Inspired by the success of deep convolutional neural networks for image classification we use a similar architecture as in \cite{krizhevsky2012imagenet} to perform end-to-end training of the CNN. \textit{TumorNet} is the CNN network trained on the lung nodule dataset. The network is comprised of 5 convolutional layers, 3 fully connected layers and a softmax classification layer. The first, second, and fifth convolutional layers are followed by a max-pooling layer. Here it is important to note that CT image patches are in 3D, whereas the inputs to the network are 2D image patches. In order to combine information across all three views of the CT, we first compute the Median Intensity Projection of the image across each view. The median projected image $\phi$ corresponding to the each dimension for an image patch $I$ is given by:
\vspace{-0.2in}
\begin{align}
\label{eqn:eqlabel}
\begin{split}
 \phi (y,z)=\median_{x} \  [I(x,y,z)], 
\\
\phi (x,z)=\median_{y} \  [I(x,y,z)], 
\\
\phi (x,y)=\median_{z} \ [I(x,y,z)],
\end{split}
\end{align}

\noindent where $\median$ is the median operator. The 3 median projected images are then concatenated to form a 3-dimensional tensor $\Phi=[\phi(y,z),\phi(x,z),\phi(x,y)]$. This tensor $\Phi$ can be considered as a 2D image with 3 channels which is used to train the CNN network. \\

\noindent \textbf{Data Augmentation:}\\
Since the number of examples is insufficient to train a deep CNN network which often required large number of training examples, we generate extra training samples from the original data. The input data is augmented using rotation and scaling. We perform random rotation of the input patch along with two different scales (one for up-sampling and the other for down-sampling). Moreover, we also add Gaussian noise with random mean, Poisson, Salt \& Pepper and Speckle noise. By applying this data augmentation strategy, we generate sufficient samples for both positive and negative examples to train our network.  

\subsection{Gaussian Process (GP) Regression:}
The deep CNN network is trained using the augmented data until the loss converges for training data. In order to extract a high-level feature representation of the input data, we use the first fully connected layer of the network to obtain a $d=4096$-dimensional feature vector. In order to reduce the computational cost, we randomly sample $n=2000$ examples from the training data and extract their corresponding features. Let $\textbf{X}=[X_1,X_2 \dots X_n]$ be the feature matrix, where $\textbf{X}\in\mathbb{R}^{n \times d}$. The regression labels are given by $\textbf{Y}=[Y_1,Y_2 \dots Y_n]$, where $\textbf{Y}\in\mathbb{R}^{n \times 1}$. For label $Y$, we use the average malignancy scores from the radiologists which lie between 1 to 5, and the objective is to regress these scores in the testing data using the regression estimator learned from the training data.

As there exists inter-observer (radiologists) variation in the malignancy scores we model it with a Gaussian Process (GP), where the prediction for an input also comes with an uncertainty measure. In our GP formulation, each feature vector $X_i$ is represented by a latent function $f_i$ with $\textbf{f}=(f_1,f_2 \dots f_n)$ which is defined as: 
\begin{equation}
\textbf{f}|\textbf{X} \sim  N(m(\textbf{X}),K(\textbf{X,X})),
\end{equation}

\noindent where $m(\textbf{X})$ is the mean function and K is the covariance matrix such that $K_{ij}=k(X_i,X_j)$. The GP regression, corresponding to a single observation $Y$ is modeled by a latent function and Gaussian noise $\epsilon$:
\begin{equation}
Y=f(X)+\epsilon, \epsilon \sim N(0,\sigma_n ^2).
\end{equation}


If $\textbf{f}$ and $\widetilde{\textbf{f}}$ represent training and testing outputs, then their joint distribution is given by:
\begin{equation}
\begin{bmatrix}
\textbf{f}\\
\widetilde{\textbf{f}} 

\end{bmatrix} \sim N\begin{pmatrix}
0, & \begin{bmatrix}
K(\textbf{X},\textbf{X}) & K(\textbf{X},\widetilde{\textbf{X}})\\ 
K(\widetilde{\textbf{X}},\textbf{X}) & K(\widetilde{\textbf{X}},\widetilde{\textbf{X}})
\end{bmatrix}
\end{pmatrix},
\end{equation}

\noindent where $K(\widetilde{\textbf{X}},\textbf{X})$ represent the covariances evaluated between all pairs of training and testing sets. Finally, the best estimator for $\widetilde{\textbf{f}}$ is computed from the mean of this distribution.
\section{Experiments and Results}
\label{sec:results}
For evaluations, we performed 10 fold cross validation over 1145 nodules. The proposed data augmentation yielded 50 extra samples corresponding to each example in the training data. We used an equal number of positive and negative examples to perform balanced training of the network without any bias. From the training set, we sampled 10\% examples to be used as validation for the CNN. The network was trained for approximately 10,000 iterations as the loss function converged around it. 

After the network was trained, we randomly sampled 2000 examples from the training data and extracted features corresponding to the first fully connected layer of the network. The GP regression was then applied to those features. The images from the test set were forward passed through the network to obtain the same feature representation followed by GP regression.

\begin{figure}[h]
\hspace{-0.15in}
\includegraphics[width=90mm,height=60mm]{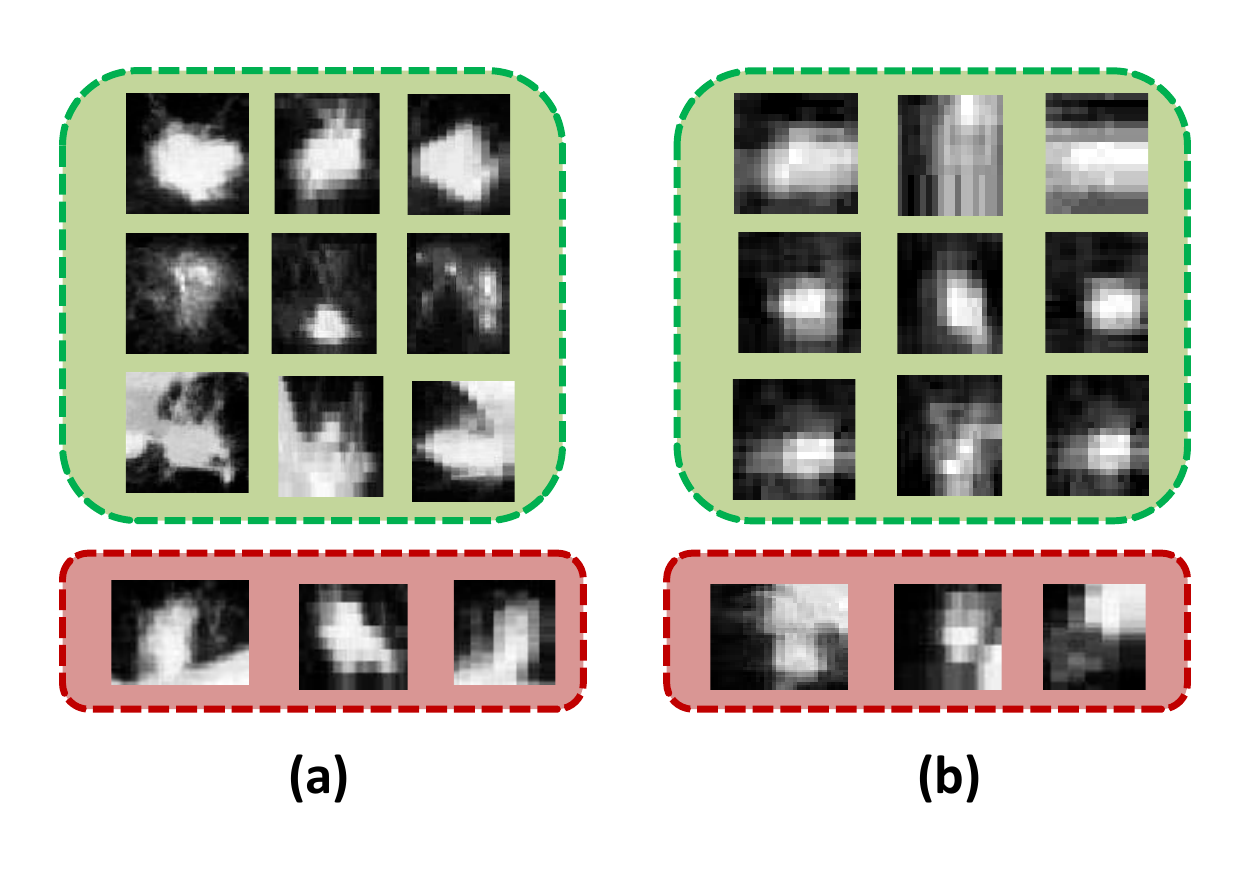}\vspace{-0.5 cm}
\caption{Qualitative results showing median intensity projected images for correctly (green) and incorrectly (red) scored lung nodules. (a) and (b) show malignant and benign nodules respectively where each row shows different cases and column represents different views (axial, sagittal, coronal).}
\label{fig:qual}
\vspace{-0.4 cm}
\end{figure}

A nodule was said to be classified successfully if its predicted score lies in $\pm1$ margin of the true score. This was done to account for any inter-observer variability in the dataset. Comparisons were performed using Support Vector Regression, Elastic Net and Least Absolute Shrinkage and Selection Operator (LASSO), where CNN features were used in all these methods. As it can be inferred from Table \ref{table:Results}, that TumorNet with GP regression outperforms popular classification and regression methods by a significant margin. Sample qualitative results are visualized in Figure \ref{fig:qual}. \\
\begin{table}[t!]
\begin{center}
\caption{\textit{Comparison of the proposed approach with Support Vector Regression, Elastic Net and LASSO using accuracy measure and standard error of the mean (SEM).
}}\label{table:Results}\vspace{-0.02in}
\label{table:quanresults_dsc}
\normalsize{
\begin{tabular}{l@{\hspace{0.10in}}c@{\hspace{0.01in}}c}
\toprule[1.5pt] \multirow{2}{*}{\textbf{Methods}}   & \multirow{2}{*}{\textbf{Regression Acc.\% (SEM\%)}} \\ 
\\
\cmidrule(r){1-3}
Support Vector Regression    &      79.91 (1.36)\    &   \\
Elastic Net       &     79.74 (0.94)     &           \\
LASSO      &     79.56 (1.14)      &           \\
\textbf{GP Regression (Proposed)}     &     \textbf{82.47 (0.62)}     &        \\
\toprule[1.5pt]
\end{tabular}}
\end{center}\vspace{-0.2in}
\end{table}

\noindent \textbf{High level Nodule Attributes:}\\
We also explored the significance of high level nodule attributes such as calcification, sphericity, texture and others for the determination of nodule malignancy. Fortunately, for the LIDC-IDRI dataset, the radiologists have also provided the scores corresponding to each of these attributes for nodules larger than 3 mm. We aim to analyze how these high level attributes can aid classification of a nodule in conjunction with the appearance features obtained using the TumorNet framework. Another reason for our interest in these high level attributes is that they can be easier to detect and annotate as compared to malignancy. In this regard, crowdsourcing can be employed to get these attributes with high efficiency and efficacy.
\begin{table}[h]
\begin{center}
\vspace{-0.3cm}
\caption{\textit{Regression accuracy and standard error (SEM) using the combination of high level attributes and CNN features.
}}\label{table:Results_att}
\label{table:quanresults_dsc}
\normalsize{
\begin{tabular}{l@{\hspace{0.05in}}c@{\hspace{0.01in}}c}
\toprule[1.5pt] \multirow{2}{*}{\textbf{Methods}}   & \multirow{2}{*}{\textbf{Regression Acc.\% (SEM\%)}} \\ 
\\
\cmidrule(r){1-3}
High level attributes   &      86.58 (0.59)     &   \\
\textbf{High level attributes+CNN}    &      \textbf{92.31 (1.59)}    &     \\
\toprule[1.5pt]
\end{tabular}}
\end{center}\vspace{-0.2in}
\end{table}

For this particular experiment, we used 6 attributes, i.e., calcification, spiculation, lobulation, margin, sphericity, and texture. We computed the average scores in cases where scores from multiple radiologists were available. We performed two sets of experiments. For first we used GP regression over the set of these 6 features and for second we concatenated them with 4096 dimension feature vector from TumorNet. We found that the combination of the high level attributes and CNN features notably improves the regression accuracy (Table \ref{table:Results_att}).

\section{Discussion and Conclusion}
\label{sec:majhead}

In contrast to the previous studies that used pre-trained network~\cite{kumar2015lung,buty2016characterization}, in this work, we proposed an end-to-end training of deep multi-view Convolutional Neural Network for nodule malignancy determination termed TumorNet. In order to cater to the need to have a large amount of labeled data for CNN, we performed data augmentation using scale, rotation and different categories of noise. In order to combine 3 different views from the image, we performed median intensity projection followed by their concatenation in a tensor form of a single image with 3 channels.

Furthermore, we explored the significance of high level nodule attributes for malignancy determination. We found that these attributes are of high significance and are actually complementary to the multi-view deep learning features. We obtained a substantial improvement in accuracy using the combination of both high level attributes and CNN features.

As an extension to this study, our future work will involve the automatic detection of high level nodule attributes and their use for malignancy determination. As these attributes may not be specific to radiology, transfer learning from other computer vision tasks can assist in addressing the challenge of the unavailability of a large amount of labeled data in radiology. Moreover, other imaging modalities such as PET could be considered for diagnostic imaging of lung cancer and treatment planning within the TumorNet framework. 
\bibliographystyle{IEEEbib}
\bibliography{egbib}

\begin{thebibliography}{10}

\bibitem{cancerstats}
N~Howlader, AM~Noone, M~Krapcho, and Garshell et~al.,
\newblock ``Seer cancer statistics review, 1975-2011,''
\newblock {\em National Cancer Institute. Bethesda, MD}, 2014.

\bibitem{uchiyama2003quantitative}
Yoshikazu Uchiyama, Shigehiko Katsuragawa, Hiroyuki Abe, Junji Shiraishi, Feng
  Li, Qiang Li, Chao-Tong Zhang, Kenji Suzuki, and Kunio Doi,
\newblock ``Quantitative computerized analysis of diffuse lung disease in
  high-resolution computed tomography,''
\newblock {\em Medical Physics}, vol. 30, no. 9, pp. 2440--2454, 2003.

\bibitem{farag2011evaluation}
Amal Farag, Asem Ali, James Graham, Aly Farag, Salwa Elshazly, and Robert Falk,
\newblock ``Evaluation of geometric feature descriptors for detection and
  classification of lung nodules in low dose ct scans of the chest,''
\newblock in {\em 2011 IEEE International Symposium on Biomedical Imaging: from
  nano to macro}. IEEE, 2011, pp. 169--172.

\bibitem{sorensen2010quantitative}
Lauge Sorensen, Saher~B Shaker, and Marleen De~Bruijne,
\newblock ``Quantitative analysis of pulmonary emphysema using local binary
  patterns,''
\newblock {\em IEEE Transactions on Medical Imaging}, vol. 29, no. 2, pp.
  559--569, 2010.

\bibitem{song2013feature}
Yang Song, Weidong Cai, Yun Zhou, and David~Dagan Feng,
\newblock ``Feature-based image patch approximation for lung tissue
  classification,''
\newblock {\em IEEE Transactions on Medical Imaging}, vol. 32, no. 4, pp.
  797--808, 2013.

\bibitem{orozco2015automated}
Hiram~Madero Orozco, Osslan Osiris~Vergara Villegas, Vianey Guadalupe~Cruz
  S{\'a}nchez, Humberto de Jes{\'u}s~Ochoa Dom{\'\i}nguez, and Manuel de
  Jes{\'u}s~Nandayapa Alfaro,
\newblock ``Automated system for lung nodules classification based on wavelet
  feature descriptor and support vector machine,''
\newblock {\em Biomedical Engineering Online}, vol. 14, no. 1, pp. 1, 2015.

\bibitem{ma2016automatic}
J~Ma, Q~Wang, Y~Ren, H~Hu, and J~Zhao,
\newblock ``Automatic lung nodule classification with radiomics approach,''
\newblock in {\em SPIE Medical Imaging}, 2016, pp. 978906--978906.

\bibitem{kumar2015lung}
Devinder Kumar, Alexander Wong, and David~A Clausi,
\newblock ``{Lung nodule classification using deep features in CT images},''
\newblock in {\em Computer and Robot Vision (CRV), 2015 12th Conference on}.
  IEEE, 2015, pp. 133--138.

\bibitem{buty2016characterization}
M~Buty, Z~Xu, M~Gao, U~Bagci, A~Wu, and D.~J Mollura,
\newblock ``Characterization of lung nodule malignancy using hybrid shape and
  appearance features,''
\newblock in {\em MICCAI}, 2016.

\bibitem{armato2011lung}
S.G Armato~III, G.~McLennan, L.~Bidaut, M.~F McNitt-Gray, C.~R Meyer, A.~P
  Reeves, B.~Zhao, D.~R Aberle, C.~I Henschke, E.~A Hoffman, et~al.,
\newblock ``{The lung image database consortium (LIDC) and image database
  resource initiative (IDRI): a completed reference database of lung nodules on
  CT scans},''
\newblock {\em Medical Physics}, vol. 38, no. 2, pp. 915--931, 2011.

\bibitem{krizhevsky2012imagenet}
Alex Krizhevsky, Ilya Sutskever, and Geoffrey~E Hinton,
\newblock ``Imagenet classification with deep convolutional neural networks,''
\newblock in {\em Advances in Neural Information Processing Systems}, 2012, pp.
  1097--1105.

\end{thebibliography}

\end{document}